%% file: main.tex
\documentclass[10pt,twocolumn,letterpaper]{article}
\input{preamble}
\usepackage[breaklinks,colorlinks,allcolors=linkblue]{hyperref}
\hypersetup{
    pdftitle={FlyVQA: Seeing Video Frame Interpolation Quality through Fly Eyes},
    pdfauthor={Baoliang CHEN, Lingyu Zhu, Zhiwei Zhong, Shiqi Wang}
}
\title{FlyVQA: Seeing Video Frame Interpolation Quality through Fly Eyes}
\author{Baoliang CHEN \quad Lingyu Zhu \quad Zhiwei Zhong \quad Shiqi Wang\\[0.5em]
    \normalsize City University of Hong Kong}
\date{}

\begin{document}
\maketitle
\input{sec/0_abstract}
\input{sec/1_intro}
\input{sec/2_related_work}
\input{sec/3_method}

\input{sec/4_experiments}
\input{sec/5_conclusion}
\FloatBarrier
{
    \small
    \raggedright
    \bibliographystyle{unsrtnat}
    \bibliography{main}
}
\end{document}

%% file: preamble.tex
\usepackage[margin=0.8in,columnsep=0.25in]{geometry}
\usepackage[T1]{fontenc}
\usepackage{lmodern}
\usepackage{microtype}
\usepackage{graphicx}
\usepackage{amsmath,amssymb}
\usepackage[dvipsnames,table]{xcolor}
\usepackage{booktabs}
\usepackage[numbers,sort&compress]{natbib}
\usepackage[hyphens]{url}
\usepackage{xurl}
\usepackage[font=small,labelfont=bf]{caption}
\usepackage{subcaption}
\usepackage[shortlabels,inline]{enumitem}
\usepackage{multirow}
\usepackage{arydshln}
\usepackage{tabularx}
\usepackage{placeins}

\definecolor{linkblue}{RGB}{32,79,128}
\definecolor{flywash}{RGB}{235,244,244}
\definecolor{tablehead}{RGB}{230,235,239}
\definecolor{tableband}{RGB}{247,249,250}
\newcommand{\msd}[2]{\ensuremath{#1{\scriptstyle\,\pm#2}}}
\newcommand{\bmsd}[2]{\ensuremath{\mathbf{#1}{\scriptstyle\,\pm#2}}}

%% file: sec/0_abstract.tex
\begin{abstract}
The ultimate ambition of visual quality assessment is to reproduce the neural computations of human vision and thereby achieve full alignment with human perception. However, the immense complexity of human visual circuitry makes this goal difficult to realize. Recently released fly visual connectomes provide a tractable biological starting point. Despite differences between fly and human vision, the fly's well-characterized motion pathways motivate us to investigate their potential for assessing the perceptual impact of temporal distortions. In this paper, we use a shared, frozen fly visual circuit to perform full-reference quality assessment of frame-interpolated videos, whose artifacts can disrupt temporal consistency. Specifically, compound-eye sampling maps reference and distorted sequences into the circuit, while a connectome-aware readout quantifies discrepancies in cell-type-specific activity and excitatory--inhibitory synaptic dynamics. Reference-guided temporal pooling and a lightweight regression head integrate these differences into a  quality score. Experiments across multiple interpolation methods demonstrate better agreement with human ratings than CNN, RNN, and Transformer baselines with comparable  parameters, supporting a biologically grounded approach to perceptual quality modeling.

\end{abstract}

%% file: sec/1_intro.tex
\section{Introduction}
\label{sec:intro}

Human perception is the ultimate reference for visual quality assessment.
Modeling the visual computations affected by distortions offers one route to perceptual alignment.
Yet human visual circuitry remains too complex and incompletely mapped to reconstruct.
Quality models therefore approximate its judgments through perceptual principles or learned representations~\cite{wang2004ssim,laparra2016nlpd,zhang2018lpips,ding2022dists}.
This leaves a fundamental question open: \textit{can an anatomically constrained visual circuit provide a useful basis for quality prediction?}

\input{figures/motivation_caption}

Fly visual connectomes make this question experimentally tractable~\cite{takemura2013motion,nern2025inventory}.
Anatomical accessibility, however, does not guarantee perceptual relevance.
Differences in retinal sampling and sensory encoding limit direct correspondence between fly and human vision.
We therefore focus on motion: the fly's ON/OFF pathways and direction-selective neurons provide well-characterized computations for processing visual change~\cite{maisak2013direction,shinomiya2019onoff}.
FlyVis combines these anatomical constraints with motion-task optimization to predict measured neural responses~\cite{lappalainen2024connectome}.
This allows us to investigate temporal quality through a circuit whose cell types, connections, and dynamics are explicitly represented.

Frame interpolation makes the temporal dimension of this problem explicit.
Frames that appear plausible in isolation can produce visibly incoherent motion during playback.
Blur, ghosting, and deformation interact with temporal inconsistencies, making quality depend on how local errors evolve across frames~\cite{danier2023bvi,hou2022vfips}.
Existing interpolation-specific metrics address this challenge through optical-flow discrepancies or learned spatiotemporal features~\cite{danier2022flolpips,hou2022vfips}.
Figure~\ref{fig:motivation} presents a biological perspective on this problem: interpolation changes not only the video frames, but also the responses they elicit in a motion-sensitive visual circuit.

Motion sensitivity alone, however, does not establish quality sensitivity.
A circuit can respond strongly to coherent motion and to distorted motion alike.
The challenge is to identify which changes in the circuit's internal responses are informative about perceptual quality.
We use cell identities, spatial organization, and signed connectivity to compare reference and distorted sequences at three levels: retinotopic neuronal activity, excitatory--inhibitory synaptic drive dynamics, and lagged temporal coupling between these drives.
These comparisons describe both the resulting activity and the inputs that generate it.
Computing discrepancies locally before aggregation retains information about where and when the responses diverge, which can be attenuated by global feature pooling.

We develop this formulation into a full-reference video quality assessment (FR-VQA) framework with a shared, frozen, motion-pretrained FlyVis circuit~\cite{lappalainen2024connectome}.
Compound-eye sampling maps reference and distorted videos into photoreceptor inputs on a common hexagonal lattice.
A connectome-guided spatiotemporal quality representation captures the resulting response discrepancies.
Reference-guided temporal pooling integrates these discrepancies, and a lightweight regression head predicts subjective quality without updating the fly circuit.
We evaluate the framework on BVI-VFI~\cite{danier2023bvi} across content-separated splits and held-out interpolation methods.
Comparisons with CNN, RNN, GRU, and Transformer baselines under comparable quality-supervised parameter budgets reveal the predictive advantages of fly-circuit priors, and targeted ablations demonstrate the quality sensitivity of our biologically grounded response representation. Our contributions are threefold:
\begin{itemize}
    \item We introduce an FR-VQA framework that transfers a motion-aware fruit-fly visual circuit to frame-interpolation quality assessment, with the connectome-constrained recurrent network frozen during  training.
    \item We propose a connectome-guided spatiotemporal quality representation that combines cell-type-specific activity, excitatory--inhibitory driving signals, and their temporal relationships, followed by  lightweight quality aggregation.
    \item We demonstrate that a frozen biological motion circuit can achieve stronger agreement with human judgments than neural baselines with comparable quality-supervised parameter counts, highlighting the potential of connectome-constrained computation for temporal quality assessment.
\end{itemize}

%% file: figures/motivation_caption.tex
\begin{figure}[t]
  \centering
  \includegraphics[width=0.85\linewidth]{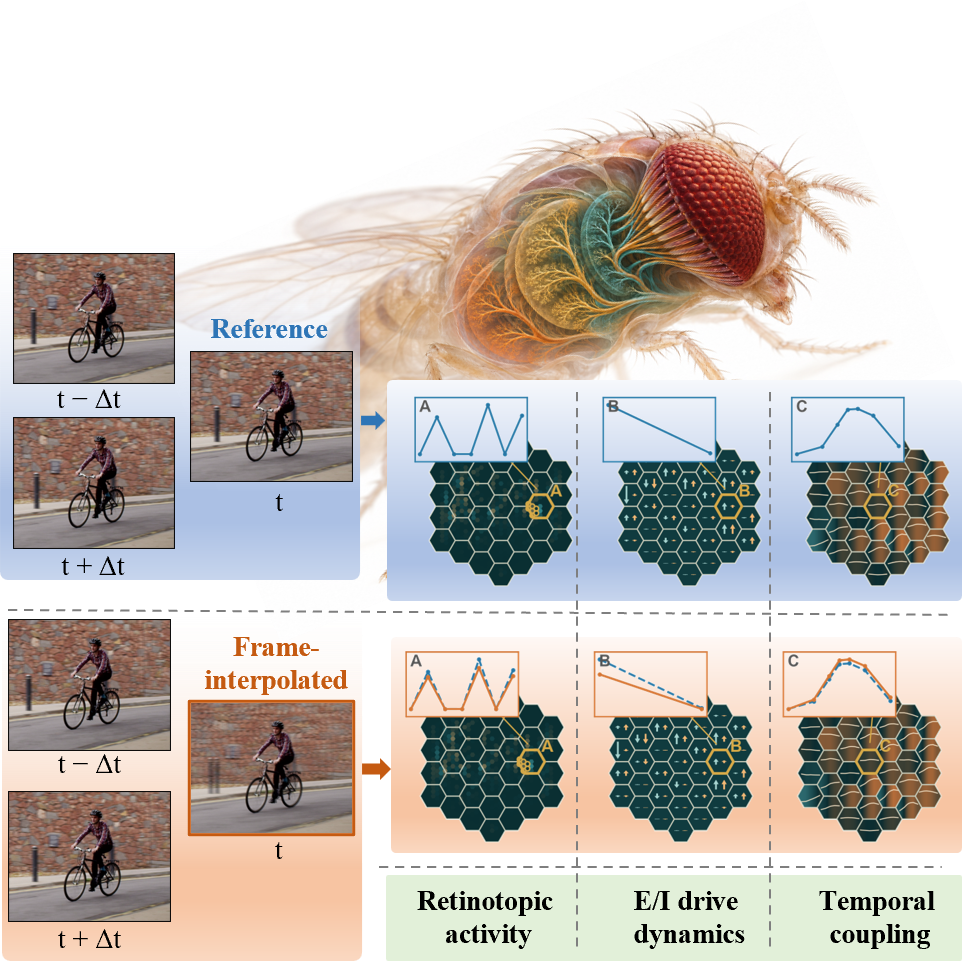}
  \caption{\textbf{Seeing temporal quality through a fly visual circuit.} An interpolated frame can appear plausible in isolation yet alter the neural responses to a moving scene. Passing reference and interpolated sequences through the same  FlyVis circuit reveals changes in retinotopic activity, excitatory--inhibitory (E/I) drive dynamics, and temporal coupling between these drives. 
  }
  \label{fig:motivation}
  \vspace{-3pt}
\end{figure}

%% file: sec/2_related_work.tex
\section{Related Work}
\label{sec:related_work}

\subsection{Video Quality Assessment}
FR-VQA evaluates spatial degradation in its temporal context.
Image metrics such as SSIM, LPIPS, and DISTS provide appearance-based comparisons~\cite{wang2004ssim,zhang2018lpips,ding2022dists}, while video-specific models account for temporal statistics and interactions.
ST-GREED measures changes in spatial and temporal band-pass entropy to predict frame-rate-dependent quality~\cite{madhusudana2021stgreed}.
C3DVQA jointly learns spatiotemporal features and quality aggregation through 2D and 3D convolutions~\cite{xu2020c3dvqa}.
Temporal aggregation also determines how localized errors contribute to a video-level judgment, making both the feature representation and its pooling relevant to quality prediction.
Frame interpolation places a particular emphasis on local motion consistency.
BVI-VFI provides a subjective benchmark across interpolation algorithms, frame rates, and resolutions~\cite{danier2023bvi}.
FloLPIPS uses optical-flow discrepancies to weight deep perceptual feature differences~\cite{danier2022flolpips}.
LPVPS, trained on the VFIPS dataset, learns interpolation-sensitive representations through feature pyramids and Swin-based spatiotemporal modules~\cite{hou2022vfips}.
These methods explicitly address temporal artifacts, motivating further investigation of the representations used to characterize them.

\subsection{Biologically Inspired Visual Modeling}
Human-vision-inspired models encode perceptual sensitivities through computational descriptions of visual processing.
The normalized Laplacian pyramid combines multiscale luminance subtraction with local gain control for perceptual image comparison~\cite{laparra2016nlpd}.
ColorVideoVDP models chromatic spatiotemporal contrast sensitivity and cross-channel masking, accounting for viewing and display conditions~\cite{mantiuk2024colorvideovdp}.
These methods demonstrate how physiological principles and psychophysical constraints can guide quality prediction.
Our investigation considers measured circuit connectivity as an additional source of biological structure.
Fly connectomics links this structure to motion computation.
Reconstructions of the ON/OFF pathways identify the synaptic organization underlying motion processing~\cite{shinomiya2019onoff}.
Mano et al. showed that anatomically constrained networks optimized for motion prediction can recover properties of individual fly neurons~\cite{mano2021anatomical}.
FlyVis combines connectome constraints with recurrent neural dynamics and optic-flow optimization to predict cell-type-specific responses~\cite{lappalainen2024connectome}.
The physiological validation of these studies concerns neuronal response properties, leaving transfer to human quality judgments as a separate empirical question.

%% file: sec/3_method.tex
\section{Method}
\label{sec:method}

\subsection{Biological Preliminaries}
\label{sec:biological_preliminaries}

\begin{figure}[t]
  \centering
  \includegraphics[width=0.95\linewidth]{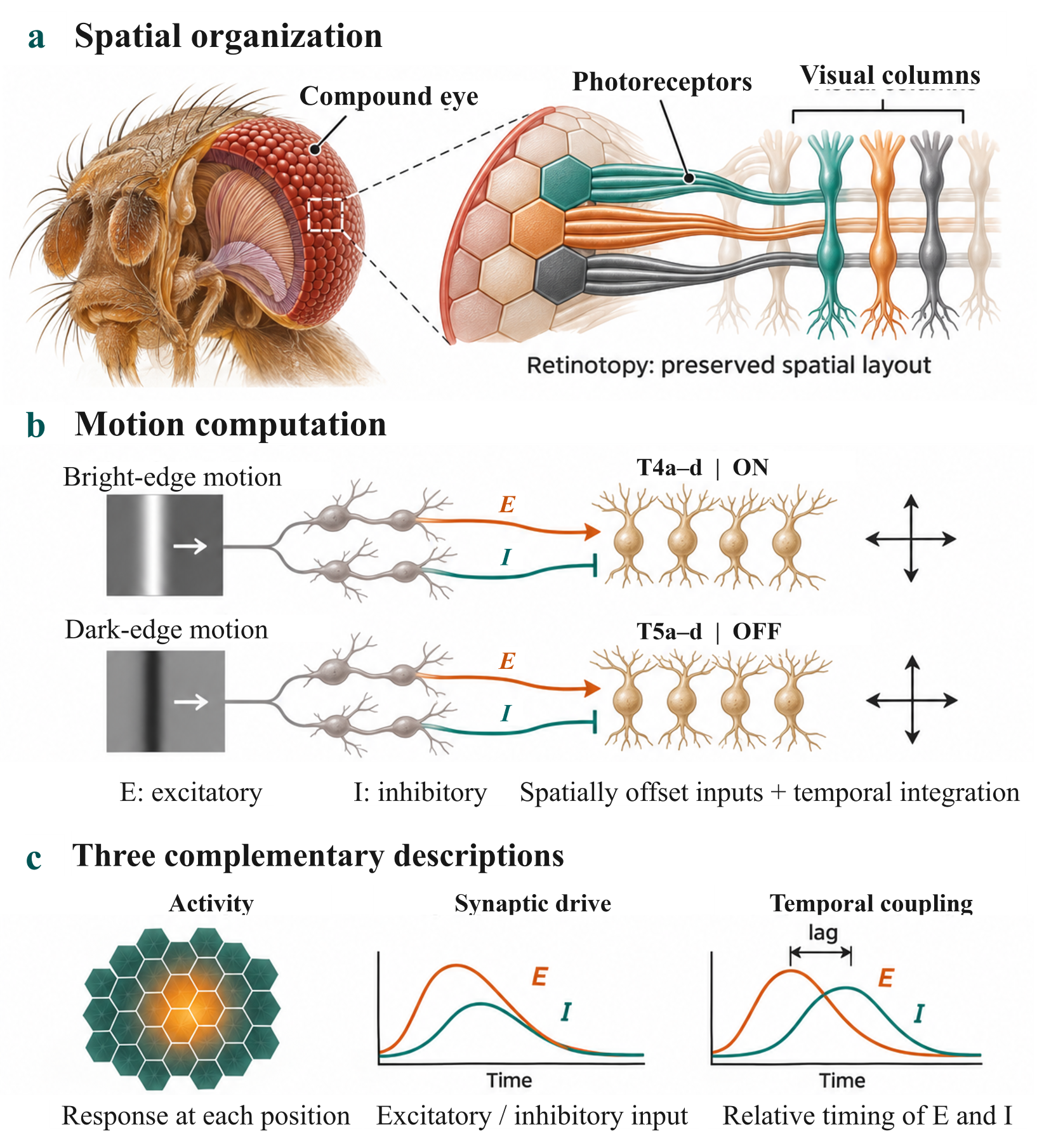}
  \caption{\textbf{Biological organization underlying our quality representation.} (a) Repeated visual columns preserve spatial correspondence with the compound eye. (b) T4 and T5 encode ON- and OFF-edge motion, respectively, with four direction-selective subtypes each. Both receive excitatory (E) and inhibitory (I) inputs. (c) This organization motivates comparing activity, synaptic drive, and their temporal relationships. }
  \label{fig:biological_preliminaries}
  \vspace{-3pt}
\end{figure}

\paragraph{From a connectome to a visual model.}
A connectome specifies which neurons are connected and the synapses between them.
FlyVis~\cite{lappalainen2024connectome} tiles local connectome reconstructions into a consensus recurrent circuit of 64 cell types.
Its nodes retain biological cell identities and spatial coordinates.
As illustrated in Fig.~\ref{fig:biological_preliminaries}a, the circuit is organized into repeated visual columns: neighboring positions in the visual field correspond to neighboring circuit locations, an organization termed \emph{retinotopy}~\cite{shinomiya2019onoff,lappalainen2024connectome}.
For a vision model, this provides spatial feature maps indexed by identified cell types rather than anonymous channels.

\paragraph{How the circuit encodes motion.}
T4 and T5 are two classes of direction-selective neurons: T4 preferentially responds to moving bright edges (ON), whereas T5 preferentially responds to moving dark edges (OFF).
Each has four subtypes, a--d, tuned to four principal motion directions~\cite{maisak2013direction}.
They receive spatially organized inputs from upstream cells~\cite{shinomiya2019onoff}.
The positive and negative connections provide excitatory (E) and inhibitory (I) drives that increase and decrease a target neuron's state, respectively.
Cell-specific time constants and recurrent interactions determine how these inputs evolve and combine over time~\cite{lappalainen2024connectome}.
ON/OFF describes stimulus contrast polarity, whereas E/I describes the sign of an input's effect; both T4 and T5 receive both kinds of drive (Fig.~\ref{fig:biological_preliminaries}b).

\paragraph{From motion responses to quality features.}
Figure~\ref{fig:biological_preliminaries}c connects this organization to three descriptions of the modeled computation: \emph{activity}, the response at each cell type and position; \emph{synaptic drive}, the excitatory and inhibitory inputs generating that response; and \emph{temporal coupling}, the relationships between these inputs at different time lags.
We compare activity across 34 output cell types, including T4/T5, and analyze drive and coupling specifically at the eight T4/T5 types.

\begin{figure*}[t]
  \centering
  \includegraphics[width=1.0\linewidth]{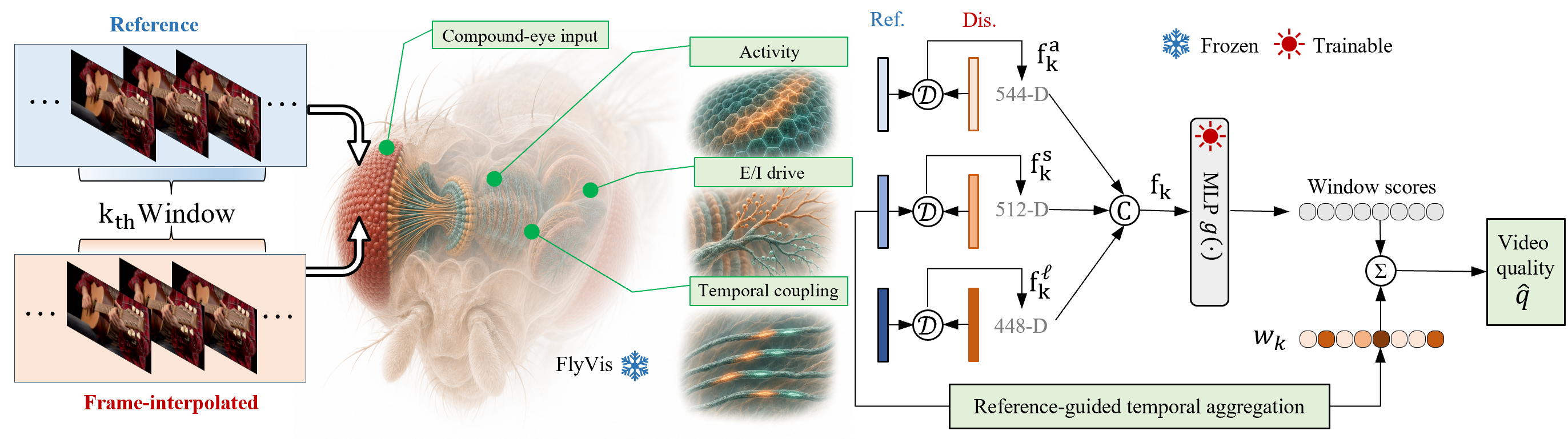}
  \caption{\textbf{Connectome-guided quality assessment of frame-interpolated videos.} Reference and distorted sequences undergo identical compound-eye sampling and independent processing by a shared, frozen FlyVis circuit. Paired activity, signed synaptic-drive, and lagged excitatory--inhibitory discrepancies form a 1504-dimensional window descriptor. A shared MLP predicts window quality before reference-response weights pool the scores over time.}
  \label{fig:framework}
  \vspace{-5pt}
\end{figure*}

\paragraph{Framework overview.}
Given a reference video $V^r$ and its frame-interpolated counterpart $V^d$, we predict a video-level subjective score $\hat q$. Figure~\ref{fig:framework} shows three components: a frozen biological motion encoder, a connectome-guided discrepancy representation, and a lightweight predictor. The shared circuit processes the two videos independently. Activity, excitatory--inhibitory (E--I) drive, and temporal-coupling discrepancies yield window descriptors. Their predicted scores are combined by reference-guided temporal weighting.

\subsection{Connectome-Guided Motion Representation}
\label{sec:motion_encoding}

\paragraph{Video-driven neural encoding.}
We use the public motion-pretrained FlyVis circuit to encode videos. Each video pair is partitioned into temporally aligned 0.5-second windows covering the entire sequence. Video frames are resized to a width of 360 pixels while preserving their aspect ratio, centered on a $391\times391$ neutral-gray canvas, and converted to luminance. The BoxEye sampler~\cite{lappalainen2024connectome} performs local spatial integration over a hexagonal lattice, producing 721 photoreceptor inputs per frame, as illustrated in Fig.~\ref{fig:biological_preliminaries}a.
The circuit dynamics are governed by the recurrent neural system
\begin{equation}
\tau_i\frac{dv_i(t)}{dt}
=
-v_i(t)+b_i+
\sum_j w_{ij}\phi(v_j(t))+s_i(t),
\label{eq:fly_dynamics}
\end{equation}
where $v_i$, $\tau_i$, and $b_i$ denote the state, time constant, and bias of neuron $i$, respectively. The signed connection weight $w_{ij}$ specifies the recurrent influence from neuron $j$ to $i$, while $\phi(v)=\max(v,0)$ defines its transmitted activity. The external stimulus $s_i$ is applied only to photoreceptor nodes.
We integrate the circuit using forward Euler updates at 240\,Hz, with a step size of $\delta t=1/240$\,s and effective time constants lower-bounded by $\delta t$. Each decoded frame is held constant until the next native frame arrives. This produces 120 neural states per window, regardless of the input frame rate.

\paragraph{Unified discrepancy characterization.}
We construct a paired neural representation from the circuit responses, as illustrated in Fig.~\ref{fig:biological_preliminaries}c. Let $z^r_{t,c,p}$ and $z^d_{t,c,p}$ denote the neural signals from the reference and distorted branches, where $t$, $c$, and $p$ index the integration step, cell type, and retinotopic position, respectively. 
We characterize temporal response variations using the scaled first-order difference
$\Delta z_t=\gamma(z_t-z_{t-1})$,
where $\gamma>0$ scales adjacent-state increments on the fixed integration grid.
To summarize neural discrepancies across space and time, we adopt a statistical operator
\begin{equation}
\mathcal{Q}_{\xi}(x)=
\left[
\operatorname{Mean}_{\xi}(x),
\operatorname{RMS}_{\xi}(x),
P_{90,\xi}(x),
\operatorname{Max}_{\xi}(x)
\right],
\label{eq:statistical_operator}
\end{equation}
where $\xi\in\{t,p\}$ specifies temporal or spatial aggregation, $\operatorname{RMS}$ denotes the root-mean-square, and $P_{90}$ is the 90th percentile. For paired neural signals, we further define a shared discrepancy operator:
\begin{equation}
\mathcal{D}(z^r,z^d)=
\mathop{\operatorname{Concat}}_{
\substack{
h\in\{\operatorname{Id},\Delta\}\\
P\in\mathcal{P}_p
}}
\mathcal{Q}_t
\left(P\left(|h(z^r)-h(z^d)|\right)\right),
\label{eq:discrepancy_operator}
\end{equation}
where
$\mathcal{P}_p=\{\operatorname{Mean}_p,\operatorname{RMS}_p\}$ and $\operatorname{Id}$ is the identity operator. Increments use valid adjacent states.
The operator computes local paired errors before spatial and temporal pooling, yielding 16 coordinates per signal type that summarize average deviations and pronounced transients. We apply these statistics to activity and synaptic drive.

\paragraph{Retinotopic activity discrepancy.}
Neural activity captures the circuit's output responses to visual stimulation. We extract the rectified activities $a^r$ and $a^d$ from 34 output cell types across all 721 retinotopic positions. Their discrepancies are characterized in terms of both response magnitudes and temporal increments through the shared operator:
\begin{equation}
\mathbf{f}^{a}_k=
\mathop{\operatorname{Concat}}_{c=1}^{34}
\mathcal{D}(a_c^r,a_c^d)
\in\mathbb{R}^{544},
\label{eq:activity_features}
\end{equation}
where $k$ indexes the temporal window.

\paragraph{Excitatory--inhibitory drive discrepancy.}
While neural activity reflects the resulting circuit responses, the signed connectome further exposes the underlying excitatory and inhibitory contributions. Specifically, the recurrent input to neuron $i$ is decomposed into
\begin{equation}
\begin{aligned}
u_i^E(t)&=\sum_j [w_{ij}]_+\phi(v_j(t)),\\
u_i^I(t)&=\sum_j [-w_{ij}]_+\phi(v_j(t)),
\end{aligned}
\label{eq:synaptic_drives}
\end{equation}
where $[x]_+=\max(x,0)$. The two terms represent nonnegative excitatory and inhibitory drive magnitudes, respectively, satisfying
$u_i^E-u_i^I=\sum_j w_{ij}\phi(v_j)$.

We extract these model-computed drives for the eight T4/T5 cell types before each state update. After subtracting the neutral-stimulus baseline, each drive is normalized by its RMS over training references, yielding the signed evoked signals $E$ and $I$. 
Negative evoked values indicate reduced drive relative to baseline, not a reversal of synaptic sign.
We construct four drive components,
$\mathcal{U}=\{E,I,E-I,E+I\}$,
which characterize excitation, inhibition, normalized E--I balance, and combined modulation, respectively. Since excitation and inhibition are normalized separately, $E-I$ describes their relative balance rather than the raw net recurrent input. Applying the shared discrepancy operator to each component yields
\begin{equation}
\mathbf{f}^{s}_k=
\mathop{\operatorname{Concat}}_{
\substack{
c=1,\ldots,8\\
z\in\mathcal{U}
}}
\mathcal{D}(z_c^r,z_c^d)
\in\mathbb{R}^{512}.
\label{eq:drive_features}
\end{equation}

\paragraph{Lagged temporal-coupling discrepancy.}
Beyond response magnitudes, the temporal coordination between excitation and inhibition provides an additional description of circuit dynamics. To capture distortion-induced changes in these interactions, we compute the lagged correlation and mean product of the E--I signals within each branch:
\begin{equation}
\begin{aligned}
C_\ell(E,I)
&=\operatorname{Corr}_{t\in\mathcal{T}_\ell}
(E_t,I_{t+\ell}),\\
M_\ell(E,I)
&=\operatorname{Mean}_{t\in\mathcal{T}_\ell}
(E_tI_{t+\ell}),
\end{aligned}
\label{eq:lagged_statistics}
\end{equation}
where $\mathcal{T}_\ell$ contains valid temporal indices for lag $\ell$. The correlation measures temporal co-variation, while the mean product additionally preserves joint amplitude information. The correlation denominator is $\sqrt{(\sigma_E^2+\eta)(\sigma_I^2+\eta)}$, with $\eta=10^{-4}$ and variances computed over the same valid index pairs. Positive lag pairs excitation with later inhibition.
We consider seven temporal offsets,
$\mathcal{L}=\{-10,-5,-2,0,2,5,10\}$,
spanning approximately $-41.7$ to $41.7$\,ms.
For each T4/T5 cell type and spatial position, we independently compute $C_\ell$ and $M_\ell$ for the reference and distorted branches. Their discrepancies are defined as
\begin{equation}
\begin{aligned}
e^C_{\ell,c,p}
&=|C^r_{\ell,c,p}-C^d_{\ell,c,p}|,\\
e^M_{\ell,c,p}
&=|M^r_{\ell,c,p}-M^d_{\ell,c,p}|.
\end{aligned}
\label{eq:coupling_errors}
\end{equation}
Unlike the previous descriptors, temporal relationships are evaluated before comparing the two branches. Since temporal information has already been summarized, we apply only spatial aggregation using $\mathcal{Q}_p$:
\begin{equation}
\mathbf{f}^{\ell}_k=
\mathop{\operatorname{Concat}}_{
\substack{
c=1,\ldots,8\\
\ell\in\mathcal{L}
}}
\left[
\mathcal{Q}_p(e^C_{\ell,c});
\mathcal{Q}_p(e^M_{\ell,c})
\right]
\in\mathbb{R}^{448}.
\label{eq:coupling_features}
\end{equation}
Finally, the three descriptors are concatenated to form a unified neural discrepancy representation for each window:
\begin{equation}
\mathbf{f}_k=
[\mathbf{f}_k^a;\mathbf{f}_k^s;\mathbf{f}_k^\ell]
\in\mathbb{R}^{1504}.
\label{eq:quality_descriptor}
\end{equation}

\subsection{Reference-Guided Quality Prediction}
\label{sec:quality_prediction}

The descriptor $\mathbf{f}_k$ is then standardized using training-set statistics and passed to $g_\theta$, a two-layer $1504\!\rightarrow\!12\!\rightarrow\!1$ MLP with SiLU activation. Only this quality regressor is trained and the FlyVis encoder remains fixed.

\paragraph{Reference-guided temporal aggregation.}
Reference-dependent weighting emphasizes windows with stronger neural dynamics while covering the full video duration. For each window $k$, we measure the temporal variation of reference T4/T5 activity as
\begin{equation}
r_k=
\operatorname{RMS}_{t,c,p}
\left(\Delta a^r_{k,t,c,p}\right),
\label{eq:reference_dynamics}
\end{equation}
where the statistic is computed over the eight T4/T5 types, all spatial positions, and valid integration steps after an initial transient.
Let $b_k$ denote the normalized duration contribution of window $k$, satisfying $\sum_k b_k=1$, with overlap accounted for when applicable. We combine duration-based and reference-response-based weighting as
\begin{equation}
\begin{aligned}
\alpha_k
&=
\frac{b_k(r_k+\epsilon)}
{\sum_m b_m(r_m+\epsilon)},\\
w_k
&=(1-\lambda)b_k+\lambda\alpha_k,
\end{aligned}
\label{eq:reference_pooling}
\end{equation}
where $\alpha_k$ assigns relatively greater importance to windows with stronger reference dynamics, while $b_k$ ensures temporal coverage. We set $\lambda=0.5$ and
$\epsilon=\max(0.1\,\operatorname{median}_{\mathrm{train}}r,10^{-8})$
to stabilize the normalization and the resulting weights satisfy $\sum_k w_k=1$.

\paragraph{Video-level prediction and optimization.}
The standardized video-level prediction and its conversion to the original DMOS scale are
\begin{equation}
\begin{aligned}
\widehat{\widetilde q}&=\sum_k w_k\,g_\theta(\widetilde{\mathbf{f}}_k),\\
\hat q&=\mu_{\mathrm{tr}}+\sigma_{\mathrm{tr}}\widehat{\widetilde q}.
\end{aligned}
\label{eq:video_quality}
\end{equation}
where $\widetilde{\mathbf{f}}_k$ denotes the standardized descriptor. 
We optimize the quality predictor using the SmoothL1 loss:
\begin{equation}
\begin{aligned}
\mathcal{L}_{\mathrm{quality}}
&=\frac{1}{B}\sum_{n=1}^{B}
\operatorname{SmoothL1}(\widehat{\widetilde q}_n,\widetilde{y}_n),\\
\widetilde{y}_n
&=\frac{y_n-\mu_{\mathrm{tr}}}{\sigma_{\mathrm{tr}}},
\end{aligned}
\label{eq:quality_loss}
\end{equation}
where $B$ is the number of videos in a batch, $y_n$ is video $n$'s differential mean opinion score (DMOS), and $\mu_{\mathrm{tr}}$ and $\sigma_{\mathrm{tr}}$ are the training-label mean and standard deviation.

%% file: sec/4_experiments.tex
\section{Experiments}
\label{sec:experiments}
\begingroup
\setlength{\textfloatsep}{6pt plus 1pt minus 1pt}
\setlength{\floatsep}{4pt plus 1pt minus 1pt}
\setlength{\dbltextfloatsep}{6pt plus 1pt minus 1pt}
\setlength{\dblfloatsep}{4pt plus 1pt minus 1pt}
\setlength{\intextsep}{6pt plus 1pt minus 1pt}
\captionsetup{skip=2pt}
\makeatletter
\renewcommand\paragraph{\@startsection{paragraph}{4}{\z@}{.9ex plus .2ex minus .1ex}{-1em}{\normalfont\normalsize\bfseries}}
\makeatother

\subsection{Experimental Setup}
\label{sec:experimental_setup}
\vspace{-2pt}

\paragraph{Dataset and training.}
BVI-VFI~\cite{danier2023bvi} contains 108 reference and 540 distorted videos from 36 contents, five interpolation methods, and three frame rates (30, 60, and 120\,Hz).
Content generalization uses ten shared splits of 24/6/6 training/validation/test contents; all versions of a content remain in one partition.
We report the mean and standard deviation of Spearman rank correlation (SRCC) and Pearson linear correlation (PLCC) against DMOS.
The former measures quality ordering and the latter linear agreement; higher values indicate better prediction.
Method generalization trains on four interpolation methods and tests on the fifth, with source contents shared.
The quality head uses AdamW~\cite{loshchilov2019decoupled} (learning rate and weight decay $10^{-3}$), batches of 32 videos, cosine decay, and gradient clipping at 1; the motion-pretrained circuit stays frozen.

\paragraph{Compared methods.}
Table~\ref{tab:overall} compares fixed metrics with retinal- and RGB-input learned baselines.
Retinal models share our preprocessing; MLP, CNN, RNN, GRU, and Transformer use 0.018--0.019M quality-supervised parameters, matching our 0.0181M readout.
The retinal CNN uses two local hexagonal convolutional layers, while the MLP operates on retinal discrepancy statistics.
RGB controls use resized color frames, with spatial models pooling frame scores and ConvRNN learning temporal features.
These conventional backbones train from scratch on identical content splits.
BoxEye RMS is our input-distance control, computed directly from paired retinal samples. Backbone citations indicate architecture origins.

\input{tables/overall}
\input{tables/cross_method}

\vspace{-3pt}
\subsection{Quality Prediction and Generalization}
\label{sec:overall_results}
\vspace{-2pt}

\paragraph{Unseen contents.}
FlyVQA leads in mean SRCC/PLCC ($0.733/0.738$), exceeding the BoxEye RNN by $0.055/0.062$ with comparable quality-supervised parameter counts.
The gain also holds over input-level controls: BoxEye RMS reaches $0.641/0.509$, and the retinal-statistics MLP reaches $0.674/0.651$.
Circuit responses thus improve on retinal-input controls.
Against the strongest fixed metrics, FlyVQA improves SRCC by $0.059$ over DMM and PLCC by $0.135$ over DISTS.
Its SRCC standard deviation is $0.033$, compared with $0.090$ for the retinal RNN and $0.110$ for the GRU, showing less variation across the shared content splits.
FlyVQA also exceeds the larger RGB backbones.

\paragraph{Unseen interpolation methods.}
\label{sec:cross_method_results}
FlyVQA obtains the highest mean SRCC ($0.613$) and a PLCC of $0.638$ in Table~\ref{tab:cross_method}, exceeding every fixed metric and learned retinal baseline in both measures.
Mean gains over learned retinal baselines are at least $0.066/0.057$.
On Repeat, it reaches $0.633/0.627$, improving on the best competing SRCC/PLCC by $0.070/0.075$.
Repetition preserves individual frames but disrupts motion continuity~\cite{danier2023bvi}, highlighting temporal sensitivity.

Average and DVF give different rankings.
Average introduces motion blur~\cite{danier2023bvi}; RGB ResNet-18 reaches $0.757/0.721$, compared with our $0.659/0.620$.
On DVF, RGB VGG-11 also achieves higher SRCC ($0.643$ vs.\ $0.527$).
Flow-based synthesis can introduce local deformation and ghosting around motion boundaries~\cite{danier2023bvi,liu2017voxel}.
Our circuit receives spatially integrated luminance on a 721-site lattice, removing color and attenuating fine texture and edge differences available to human observers.
This reduced spatial detail plausibly contributes to the RGB advantage, motivating complementary appearance cues.

\begin{figure*}[!t]
  \centering
  \includegraphics[width=0.8\linewidth]{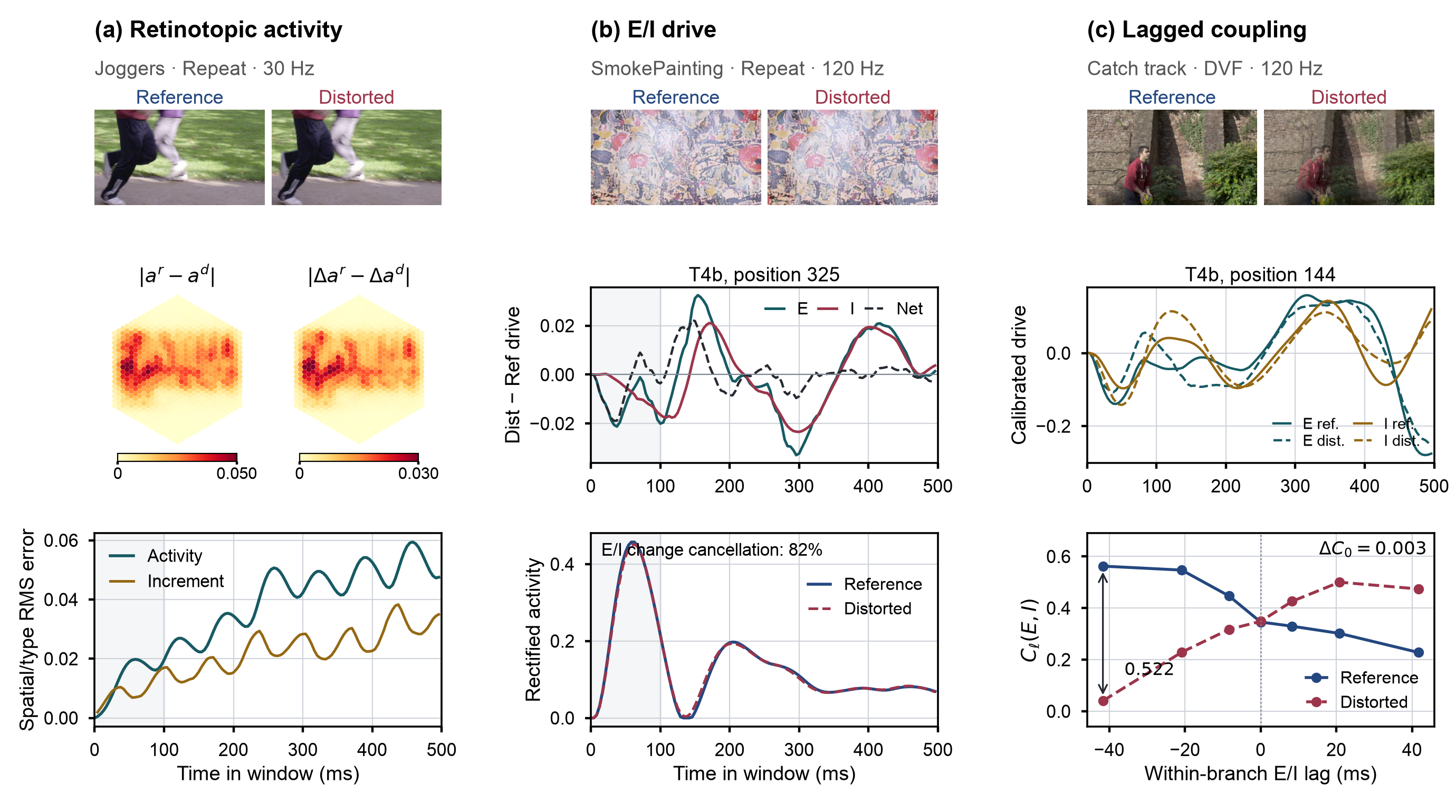}
  \vspace{-2pt}
  \caption{\textbf{Circuit responses to interpolation distortions.}
  Three 0.5\,s windows: \textbf{(a)} activity/increment errors (Joggers, Repeat, 30\,Hz); \textbf{(b)} E/I changes with similar activity (SmokePainting, Repeat, 120\,Hz); \textbf{(c)} lagged coupling changes (Catch track, DVF, 120\,Hz).
  Maps are averaged over time and 34 cell types; (b) uses raw drive changes and (c) training-scaled evoked drives. }
  \label{fig:response_cases}

\end{figure*}

\vspace{-3pt}
\subsection{Distortion Signatures in Circuit Responses}
\label{sec:response_analysis}
\vspace{-2pt}

We probe the frozen circuit at three levels: spatially organized output activity, incoming E/I drives, and their temporal coordination.
Reference and distorted inputs use identical retinal sampling and 240\,Hz integration.
Each analysis window lasts 0.5\,s and contains 120 neural states.

\paragraph{Localized activity discrepancies.}
For the Joggers/Repeat example, we compare activity and temporal increments at corresponding retinal positions, then average absolute errors over time and 34 output types to obtain the two maps in Fig.~\ref{fig:response_cases}a.
RMS pooling over positions and types at each time step produces the lower traces.
The maps localize discrepancies; traces separate activity and increment errors.
To verify localization under a known perturbation, we blur a fixed image region before retinal sampling and run the unchanged circuit.
Mean activity error is $18.2\times$ larger inside the affected retinal footprint than outside, evaluated over positions with varying reference responses.
Thus, local response comparisons preserve evidence that whole-field averaging can dilute.

\paragraph{Drive changes beneath similar activity.}
In the SmokePainting/Repeat window, we inspect an active T4b neuron and compare distorted-minus-reference excitatory drive, inhibitory drive, their signed net input, and output activity (Fig.~\ref{fig:response_cases}b).
Similar E/I changes can offset because their signs oppose.
Cancellation is measured in raw drive units as one minus the integrated absolute net-drive change divided by the integrated sum of absolute E/I changes; separate E/I normalization is used only for the feature descriptors.
Over 100--500\,ms, $82.2\%$ cancels, leaving a mean  activity difference of only $0.0032$ against a reference temporal standard deviation of $0.051$.
Comparing E and I separately therefore exposes changes that nearly disappear at the output, giving the drive descriptors a distinct role alongside activity.

\paragraph{Temporal coordination beyond amplitude.}
For the Catch track/DVF window, we compute within-branch E/I correlations at seven fixed lags before comparing the two branches (Fig.~\ref{fig:response_cases}c).
At the illustrated T4b neuron, distorted/reference E/I RMS ratios are $0.93/0.95$, and the zero-lag correlation changes by only $0.003$.
The discrepancy reaches $0.522$ at $-41.7$\,ms.
It persists after matching individual E/I means and variances; excluding the first 100\,ms still leaves a maximum lagged discrepancy of $0.448$.
Lag profiles therefore expose coordination changes missed by amplitude and zero-lag statistics.

\begin{figure}[!t]
  \centering
  \includegraphics[width=\linewidth]{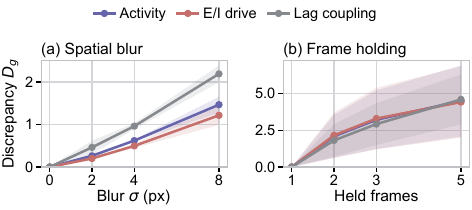}
  \vspace{-5pt}
  \caption{\textbf{Controlled spatial and temporal sensitivity.}
  Feature-group magnitudes $D_g$ under blur and frame holding.
}
  \label{fig:response_validation}

\end{figure}


\paragraph{Sensitivity to controlled degradation.}
We use two 30\,Hz reference windows and perturb only the distorted branch.
Gaussian blur is applied after width-360 resizing and before retinal sampling, with $\sigma\in\{0,2,4,8\}$ pixels.
Temporal holding repeats each selected frame for $h\in\{1,2,3,5\}$ native-frame positions, preserving the frame rate and circuit integration step.
Here $h=1$ is unchanged; larger $h$ prolongs frame persistence without changing duration.
For feature group $g$ of dimension $d_g$, we compute
\begin{equation}
D_g(k)=\sqrt{\frac{1}{d_g}\sum_{j\in g}\left(\frac{f_{k,j}}{s_j}\right)^2},
\label{eq:diagnostic_magnitude}
\end{equation}
where $s_j$ is coordinate $j$'s RMS over training windows, with a numerical floor.
All three groups increase monotonically with both perturbations in each window (Fig.~\ref{fig:response_validation}), starting from zero discrepancy for the unmodified references.
The circuit responds to both spatial softening and interrupted temporal evolution, linking stronger degradation to larger response discrepancies.

\paragraph{Association with subjective impairment.}
We next analyze 90 held-out videos: six contents, three frame rates, and five interpolation methods.
Each video supplies ten 0.5\,s windows. Duration pooling converts their $D_g$ values into one magnitude per feature group.
Within each of the 18 content/frame-rate groups, we subtract the five-method mean from both magnitudes and DMOS, then compute SRCC over the 90 centered pairs.
Centering removes the shared content/frame-rate offset, so the analysis compares method-dependent impairment rather than overall differences between scenes.
Activity, drive, and coupling reach $0.552$, $0.568$, and $0.552$ SRCC, respectively, connecting their response sensitivity to human judgments without fitting a quality regressor for this analysis.

\subsection{Ablation Studies}
\label{sec:ablations}
\vspace{-2pt}

We examine whether fly-circuit priors offer advantages beyond generic neural capacity and  pretraining, and whether our biological response representation captures perceptually relevant distortions. 
\input{tables/ablation}
\input{tables/ablation_auxiliary}

\paragraph{Circuit priors.}
Table~\ref{tab:circuit_priors}a compares encoders  with comparable  parameter budgets.
Without motion pretraining, frozen FlyVis with random strengths reaches $0.696/0.661$, exceeding scratch-trained RNN and GRU in SRCC.
This ordering supports useful inductive biases from the circuit and its structured representation.
Motion pretraining further raises frozen FlyVis to $0.733/0.738$.
Panel~(b) evaluates motion transfer: the original RNN and GRU  are pretrained  on Sintel optical flow~\cite{butler2012sintel}, using the same dataset and prediction task as FlyVis, then frozen while fresh quality heads are trained.
They reach $0.508/0.429$ and $0.672/0.625$, below FlyVQA's $0.733/0.738$.
Permuting trained strength scales within E/I groups, while fixing wiring, signs, synapse counts, biases, and time constants, reduces agreement to $0.690/0.662$.
These results support the circuit-informed representation and learned strength assignment.

\begin{figure}[!t]
  \centering
  \includegraphics[width=\linewidth]{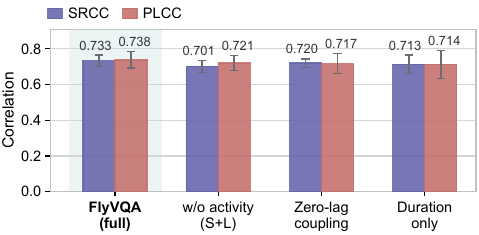}
  \vspace{-5pt}
  \caption{\textbf{Contributions of representations.}
  All variants use the same predictor, and removed standardized coordinates are zeroed.}
  \label{fig:ablation}
  \vspace{-5pt}
\end{figure}

\begin{figure}[t]
    \centering
    \includegraphics[width=\linewidth]{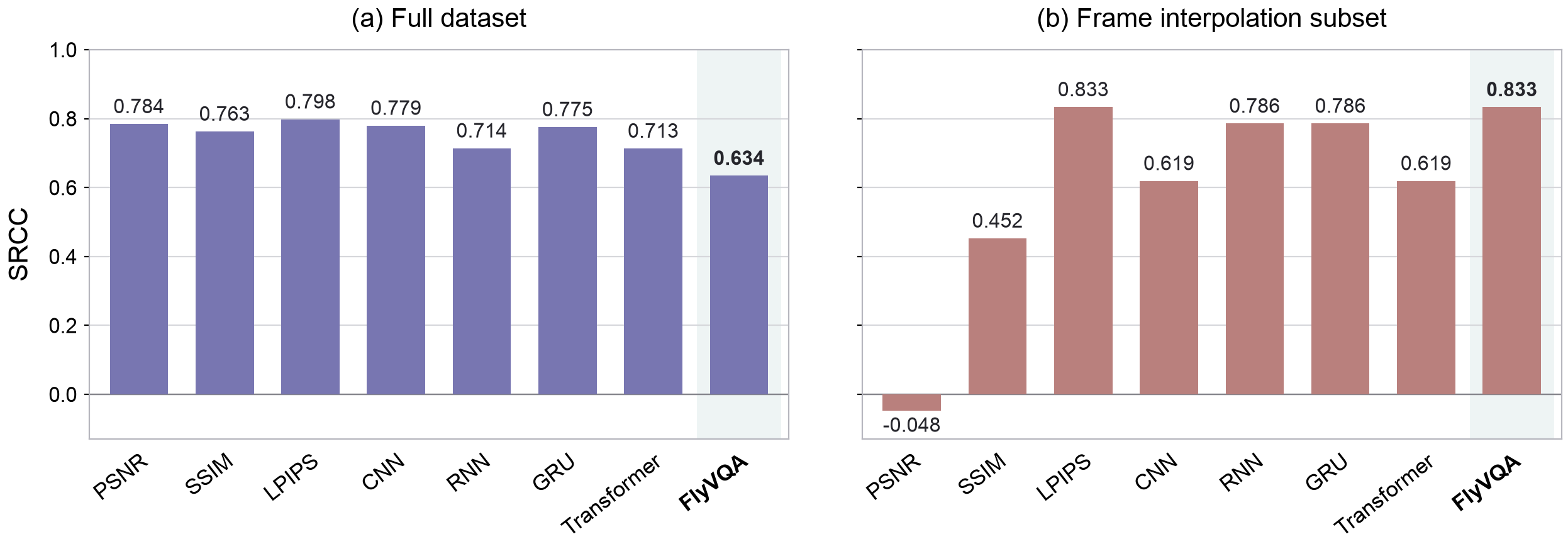}
    \vspace{-5pt}
    \caption{\textbf{Cross-dataset evaluation on CGVQD.}}
    \label{fig:cgvqd_transfer}
    \vspace{-5pt}
\end{figure}

\paragraph{Activity and feature combinations.}
Table~\ref{tab:feature_combinations} presents the feature-combination study.
Activity raises SRCC from $0.674$ for input statistics to $0.725$.
A+S yields a higher SRCC, whereas A+S+L gives the highest, showing different ranking agreement tradeoffs rather than uniform gains from every addition.
Under the final  protocol, removing activity while retaining S+L lowers SRCC from $0.733$ to $0.701$ (Fig.~\ref{fig:ablation}), with lower SRCC on all ten splits.
Output activity thus complements internal synaptic dynamics.

\paragraph{Nonzero-lag coupling.}
The zero-lag control preserves activity, drive, and both E/I correlation and mean product at lag zero, removing only the six nonzero-lag terms.
SRCC/PLCC fall to $0.720/0.717$ (Fig.~\ref{fig:ablation}), supporting temporal coordination beyond simultaneous co-variation.

\paragraph{Reference-guided temporal aggregation.}
With all features maintained, we replace reference-response weights by duration weights throughout training  and testing, and keep the same predictor from original initialization.
Agreement decreases to $0.713/0.714$ (Fig.~\ref{fig:ablation}).
The full rule improves mean SRCC/PLCC by $0.020/0.024$, supporting reference dynamics for learning and aggregating window predictions.

\endgroup

\subsection{Limitation}
\label{sec:limitations}
\vspace{-2pt}
To assess transfer beyond interpolation artifacts, we evaluate
FlyVQA on CGVQD~\cite{jindal2025cgvqm} without adaptation
(Fig.~\ref{fig:cgvqd_transfer}). On all videos, it achieves
0.634 SRCC, below LPIPS and BoxEye CNN/GRU, while interpolation SRCC reaches 0.833. This gap may
reflect the specialization of our frozen motion circuit and
fixed response features. Their sensitivity to interpolation
artifacts may not extend to consistent quality rankings across
heterogeneous distortions. Combining these biological motion responses with learnable
spatial features may improve generalization across distortion types.

%% file: tables/overall.tex
\begin{table}[t]
\centering
\caption{\textbf{Quality prediction on unseen BVI-VFI content.}
Mean$\pm$standard deviation over ten splits; $P$: quality-supervised parameters (M).
$\dagger$: native-resolution VMAF.}
\label{tab:overall}
\vspace{-1pt}
\fontsize{8}{9.3}\selectfont
\setlength{\tabcolsep}{2.4pt}
\renewcommand{\arraystretch}{1.02}
\begin{tabularx}{\linewidth}{@{}Xrcc@{}}
\toprule
Method & $P$ & SRCC$\uparrow$ & PLCC$\uparrow$\\
\midrule
\rowcolor{tablehead}\multicolumn{4}{l}{\textit{Fixed quality metrics}}\\
PSNR & -- & \msd{0.658}{0.078} & \msd{0.579}{0.069}\\
SSIM~\cite{wang2004ssim} & -- & \msd{0.632}{0.093} & \msd{0.489}{0.120}\\
MS-SSIM~\cite{wang2003msssim} & -- & \msd{0.653}{0.072} & \msd{0.492}{0.124}\\
LPIPS~\cite{zhang2018lpips} & -- & \msd{0.654}{0.087} & \msd{0.523}{0.125}\\
DISTS~\cite{ding2022dists} & -- & \msd{0.655}{0.051} & \msd{0.603}{0.079}\\
DMM~\cite{chen2025debiased} & -- & \msd{0.674}{0.041} & \msd{0.525}{0.083}\\
FloLPIPS~\cite{danier2022flolpips} & -- & \msd{0.655}{0.066} & \msd{0.552}{0.087}\\
VMAF$^{\dagger}$~\cite{netflixvmaf} & -- & \msd{0.601}{0.103} & \msd{0.536}{0.108}\\
BoxEye RMS (control) & -- & \msd{0.641}{0.032} & \msd{0.509}{0.073}\\
\hdashline
\rowcolor{tablehead}\multicolumn{4}{l}{\textit{BoxEye-input FR-VQA}}\\
MLP & 0.0180 & \msd{0.674}{0.053} & \msd{0.651}{0.065}\\
CNN & 0.0190 & \msd{0.672}{0.050} & \msd{0.632}{0.073}\\
RNN~\cite{elman1990finding} & 0.0187 & \msd{0.678}{0.090} & \msd{0.676}{0.074}\\
GRU~\cite{cho2014learning} & 0.0190 & \msd{0.655}{0.110} & \msd{0.637}{0.081}\\
Transformer~\cite{vaswani2017attention} & 0.0186 & \msd{0.669}{0.108} & \msd{0.676}{0.077}\\
ResNet-18~\cite{he2016deep} & 11.308 & \msd{0.615}{0.059} & \msd{0.529}{0.086}\\
\hdashline
\rowcolor{tablehead}\multicolumn{4}{l}{\textit{RGB-input FR-VQA}}\\
ResNet-18~\cite{he2016deep} & 11.308 & \msd{0.615}{0.073} & \msd{0.529}{0.109}\\
VGG-11~\cite{simonyan2015very} & 9.352 & \msd{0.659}{0.100} & \msd{0.602}{0.134}\\
Spatial ViT~\cite{dosovitskiy2021image} & 2.924 & \msd{0.593}{0.095} & \msd{0.468}{0.138}\\
ConvRNN & 0.0422 & \msd{0.662}{0.050} & \msd{0.587}{0.081}\\
\midrule
\rowcolor{flywash}\textbf{FlyVQA (Ours)} & \shortstack[r]{0.0181\\(total 0.0218)} & \bmsd{0.733}{0.033} & \bmsd{0.738}{0.046}\\
\bottomrule
\end{tabularx}
\vspace{-2pt}
\end{table}

%% file: tables/cross_method.tex
\begin{table*}[t]
\centering
\caption{\textbf{Generalization to unseen interpolation methods.}
Learned models train on four methods and test on the fifth, with source contents shared.
Values are SRCC / PLCC; Mean averages the five methods.}
\label{tab:cross_method}
\vspace{-1pt}
\fontsize{8.1}{9.5}\selectfont
\setlength{\tabcolsep}{3.1pt}
\renewcommand{\arraystretch}{1.02}
\begin{tabularx}{\textwidth}{@{}l*{6}{>{\centering\arraybackslash}X}@{}}
\toprule
Method & Repeat & Average & DVF~\cite{liu2017voxel} & QVI~\cite{xu2019quadratic} & ST-MFNet~\cite{danier2022stmfnet} & Mean\\
\midrule
\rowcolor{tablehead}\multicolumn{7}{l}{\textit{Fixed quality metrics}}\\
BoxEye RMS (control) & 0.389 / 0.338 & 0.493 / 0.420 & 0.377 / 0.347 & 0.685 / 0.655 & \textbf{0.623} / 0.723 & 0.513 / 0.496\\
DISTS~\cite{ding2022dists} & 0.356 / 0.326 & 0.520 / 0.525 & 0.334 / 0.394 & 0.693 / 0.725 & 0.592 / 0.720 & 0.499 / 0.538\\
DMM~\cite{chen2025debiased} & 0.363 / 0.234 & 0.513 / 0.438 & 0.363 / 0.310 & 0.690 / 0.599 & 0.611 / 0.738 & 0.508 / 0.464\\
FloLPIPS~\cite{danier2022flolpips} & 0.398 / 0.312 & 0.495 / 0.449 & 0.290 / 0.309 & 0.662 / 0.722 & 0.588 / 0.718 & 0.487 / 0.502\\
\hdashline
\rowcolor{tablehead}\multicolumn{7}{l}{\textit{BoxEye-input FR-VQA}}\\
MLP & 0.447 / 0.469 & 0.622 / 0.616 & 0.493 / 0.462 & 0.684 / 0.714 & 0.491 / 0.642 & 0.547 / 0.581\\
CNN & 0.402 / 0.368 & 0.604 / 0.511 & 0.400 / 0.411 & 0.684 / \textbf{0.758} & 0.537 / 0.655 & 0.525 / 0.541\\
RNN~\cite{elman1990finding} & 0.403 / 0.392 & 0.542 / 0.512 & 0.326 / 0.379 & 0.702 / 0.677 & 0.590 / 0.654 & 0.513 / 0.523\\
GRU~\cite{cho2014learning} & 0.446 / 0.443 & 0.544 / 0.503 & 0.319 / 0.380 & 0.689 / 0.694 & 0.568 / 0.716 & 0.513 / 0.547\\
Transformer~\cite{vaswani2017attention} & 0.410 / 0.389 & 0.482 / 0.511 & 0.422 / 0.467 & \textbf{0.724} / 0.719 & 0.527 / 0.670 & 0.513 / 0.551\\
\hdashline
\rowcolor{tablehead}\multicolumn{7}{l}{\textit{RGB-input FR-VQA}}\\
ResNet-18~\cite{he2016deep} & 0.395 / 0.356 & \textbf{0.757} / \textbf{0.721} & 0.583 / 0.635 & 0.674 / 0.604 & 0.606 / 0.692 & 0.603 / 0.602\\
VGG-11~\cite{simonyan2015very} & 0.238 / 0.184 & 0.723 / 0.697 & \textbf{0.643} / 0.631 & 0.669 / 0.512 & 0.572 / 0.589 & 0.569 / 0.523\\
Spatial ViT~\cite{dosovitskiy2021image} & 0.563 / 0.552 & 0.701 / 0.692 & 0.638 / 0.657 & 0.600 / 0.648 & 0.546 / 0.678 & 0.610 / \textbf{0.645}\\
ConvRNN & 0.390 / 0.383 & 0.713 / 0.685 & 0.608 / \textbf{0.662} & 0.617 / 0.688 & 0.562 / \textbf{0.755} & 0.578 / 0.635\\
\midrule
\rowcolor{flywash}\textbf{FlyVQA (ours)} & \textbf{0.633} / \textbf{0.627} & 0.659 / 0.620 & 0.527 / 0.569 & 0.680 / 0.685 & 0.565 / 0.687 & \textbf{0.613} / 0.638\\
\bottomrule
\end{tabularx}
\vspace{-4pt}
\end{table*}

%% file: tables/ablation.tex
\begin{table}[!t]
\centering
\caption{\textbf{Circuit priors and motion pretraining.}
Mean$\pm$sample standard deviation over ten shared splits.
(a)  E2E updates the encoder and head, while frozen updates only the head.
(b) Quality-supervised budgets are approximately 18--19k parameters.}
\label{tab:circuit_priors}
\vspace{-1pt}
\fontsize{7.8}{8.8}\selectfont
\setlength{\tabcolsep}{3pt}
\renewcommand{\arraystretch}{1.01}
\begin{tabularx}{\linewidth}{@{}Xcc@{}}
\toprule
Configuration & SRCC$\uparrow$ & PLCC$\uparrow$\\
\midrule
\rowcolor{tablehead}\multicolumn{3}{l}{\textit{(a) Circuit priors and initialization }}\\
RNN, scratch (E2E) & \msd{0.678}{0.090} & \msd{0.676}{0.074}\\
GRU, scratch (E2E) & \msd{0.655}{0.110} & \msd{0.637}{0.081}\\
FlyVis, random (frozen) & \msd{0.696}{0.035} & \msd{0.661}{0.066}\\
FlyVis, default (frozen) & \msd{0.701}{0.038} & \msd{0.708}{0.074}\\

\hdashline\noalign{\vskip1pt}
\rowcolor{tablehead}\multicolumn{3}{l}{\textit{(b) Motion-transfer controls }}\\
RNN, motion (frozen) & \msd{0.508}{0.112} & \msd{0.429}{0.134}\\
GRU, motion (frozen) & \msd{0.672}{0.046} & \msd{0.625}{0.080}\\
FlyVis, shuffled strengths & \msd{0.690}{0.038} & \msd{0.662}{0.055}\\
\rowcolor{flywash}\textbf{FlyVQA (final)} & \bmsd{0.733}{0.033} & \bmsd{0.738}{0.046}\\
\bottomrule
\end{tabularx}

\end{table}

%% file: tables/ablation_auxiliary.tex
\begin{table}[!t]
\centering
\caption{\textbf{Feature combinations.}
Mean$\pm$sample standard deviation over ten shared splits.
A/S/L denote activity/synaptic drive/lagged coupling; $D$ is the feature dimension.}
\label{tab:feature_combinations}

\fontsize{7.8}{8.8}\selectfont
\setlength{\tabcolsep}{3pt}
\renewcommand{\arraystretch}{1.01}
\begin{tabularx}{\linewidth}{@{}Xrcc@{}}
\toprule
Features & $D$ & SRCC$\uparrow$ & PLCC$\uparrow$\\
\midrule
Input statistics & 544 & \msd{0.674}{0.053} & \msd{0.651}{0.065}\\
A & 544 & \msd{0.725}{0.017} & \msd{0.710}{0.056}\\
A+S & 1056 & \bmsd{0.728}{0.025} & \msd{0.689}{0.068}\\
A+L & 992 & \msd{0.724}{0.028} & \msd{0.720}{0.056}\\
\rowcolor{flywash}A+S+L (FlyVQA) & 1504 & \bmsd{0.733}{0.033} & \bmsd{0.738}{0.046}\\

\bottomrule
\end{tabularx}

\end{table}

%% file: sec/5_conclusion.tex
\vspace{-7pt}
\section{Conclusion}
\label{sec:conclusion}

We introduced FlyVQA for frame-interpolation VQA using a frozen, motion-pretrained fly circuit.
Activity and signed synaptic-dynamics features enable lightweight quality prediction.
On BVI-VFI, the framework improves average agreement with human ratings over the evaluated baselines with comparable quality-supervised parameter counts.
These findings support biological motion representations for perceptual video temporal quality modeling.